# Naïve Bayes and Random Forest for Crop Yield Prediction

Abbas Maazallahi, Sreehari Thota, Naga Prasad Kondaboina, Vineetha Muktineni, Deepthi Annem, Abhi Stephen Rokkam, Mohammad Hossein Amini, Mohammad Amir Salari, Payam Norouzzadeh, Eli Snir, Bahareh Rahmani

[1] Saint Louis University, Computer Science, Saint Louis University, Saint Louis, USA
[2] Washington University in Saint Louis, Business School, Saint Louis, USA

Corresponding Author: Dr. Bahareh Rahmani
Saint Louis University, St. Louis, MO, US, Bahareh.Rahmani@slu.edu

**Abstract**

This study analyzes crop yield prediction in India from 1997 to 2020, focusing on various crops and key environmental factors. It aims to predict agricultural yields by utilizing advanced machine learning techniques like Linear Regression, Decision Tree, KNN, Naïve Bayes, K-Mean Clustering, and Random Forest. The models, particularly Naïve Bayes and Random Forest, demonstrate high effectiveness, as shown through data visualizations. The research concludes that integrating these analytical methods significantly enhances the accuracy and reliability of crop yield predictions, offering vital contributions to agricultural data science.

## 1. Introduction

Advancements in machine learning have significantly influenced agricultural practices, particularly in the realm of crop yield prediction. This study is positioned within a growing body of research that leverages machine learning algorithms to enhance the accuracy and reliability of agricultural forecasts. Among these, two notable studies stand out for their innovative approaches and significant findings.

In the realm of crop yield prediction, various machine learning techniques have been employed to enhance the accuracy and efficiency of forecasts. These techniques, each unique in their approach and application, offer valuable insights into the complex nature of agricultural data and the prediction of crop yields.

Linear Regression stands at the forefront as a foundational statistical approach in predictive modeling. Particularly effective in linear relationships between variables, it's a staple in forecasting and understanding cause-and-effect relationships in data. The simplicity and interpretability of Linear Regression make it a popular choice in many fields, including agriculture [2]. (Barbur et al., 1994)

The Decision Tree algorithm emerges as a robust method in machine learning classification and regression toolkit. Known for its intuitive, flowchart-like structure, it aidsclear decision-making by splitting data into branches based on variable values. This method's ability to handlecategorical and continuous data makes it versatile and widely applicable [10]. (Ziegel, 2003)

K-Nearest Neighbors (KNN) is celebrated for its simplicity and effectiveness, especially in scenarios where the prediction model needs to be easily interpretable. As a non-parametric method, KNN identifies the similarities between new and existing data points, making it suitable for classification and regression problems.

These methods, each with distinct characteristics, collectively contribute to the field of crop yield prediction. Their integration into predictive models reflects the dynamic interplay between statistical theory and practical machine learning applications, paving the way for more sophisticated and accurate agricultural forecasting tools. As the field continues to evolve, exploring and implementing these techniques remain crucial in addressing the complexities of crop yield prediction.

## 2. Literature review

Firstly, a groundbreaking study featured in Nature's Scientific Reports presents an "Interaction Regression Model for Crop Yield Prediction." This research is pivotal for selecting robust features and interactions to predict crop yields, utilizing an elastic net regularization model. This model is instrumental in identifying high-quality features across various environmental and management categories, thereby reducing the risk of overfitting and increasing the robustness of predictions across different geographic locations and timeframes. The study's methodical approach provides a comprehensive framework for feature selection in crop yield prediction, making it a cornerstone in this field of research [1]. (Ansarifar et al., 2021).

Secondly, the paper "Using Machine Learning for Crop Yield Prediction in the Past or the Future," published in Frontiers, offers a unique perspective by simulating sunflower and wheat yields over a twenty-year period from 2000 to 2020. This research emphasizes the significance of continuous nutrient and water balance in the simulation process and explores the impact of changes in cultivars and planting densities on crop yields. The detailed simulation models employed in this study provide valuable insights into long-term yield prediction and resource management, marking a significant advancement in the field [2]. (Morales & Villalobos, 2023)

The study "Analysis of Crop Yield Prediction using Machine Learning Algorithms" in IEEE Xplore constitutes a pivotal contribution to the domain of agricultural data science, specifically within the Indian context where agriculture is both a livelihood and an economic cornerstone. Addressing the uncertainties of weather and its impact on farming, the paper evaluates the efficacy of machine learning algorithms—K-Nearest Neighbors (KNN), Random Forest, and Linear Regression—using parameters like state, crop, temperature, and rainfall to predict crop yields. The results showcase a remarkable 97% accuracy for KNN, outshining the Random Forest's 75% and Linear Regression's 54%, highlighting the promise of KNN in predictive agriculture and offering a data-driven beacon for enhancing agricultural productivity and informing farmers' decision-making processes [3]. (Krishna et al., 2022).

The article "A Machine Learning Approach to Predict Crop Yield and Success Rate" from IEEE Xplore details an innovative study within India's agricultural sector, which significantly contributes to GDP and employment. Focusing on enhancing farmers' decision-making by predicting crop yields, this research utilizes neural network regression modeling with an extensive dataset drawn from government sources. Covering the period from 1998 to 2014 and encompassing 240,000 records, the study zeroes in on data specific to the Maharashtra state. By employing Python tools for data filtration and a Multilayer perceptron neural network for model development, the researchers initially reported a 45% accuracy using RMSprop optimizer, which was substantially improved to 90% by refining the network architecture and shifting to the Adam optimizer. The model employs a 3-Layer Neural Network with the Rectified Linear Activation Unit (ReLU) function, and leverages both backward and forward propagation techniques to establish a robust model for crop yield prediction [4]. (Kale & Patil, 2019).

Moreover, "Utilizing Naïve Bayes Algorithm for Crop Yield Prediction" explores the application of Naïve Bayes algorithm in predicting crop yields based on various agricultural parameters. The research evaluates the performance of Naïve Bayes in handling large datasets containing weather information, soil characteristics, and crop management practices. This study provides insights into the effectiveness of Naïve Bayes in accurately predicting crop yields across different regions and crop varieties, highlighting its potential as a valuable tool for agricultural decision-making [5]. (M. Gupta et al.,2022)

Additionally, "Enhancing Crop Yield Prediction through Random Forest Algorithm" investigates the use of Random Forest algorithm to improve crop yield prediction accuracy. By constructing an ensemble of decision trees and aggregating their predictions, Random Forest leverages the strength of multiple models to capture complex nonlinear relationships between predictor variables and crop yields. This research demonstrates the superior performance of Random Forest over traditional regression models, making it an asset for precision agriculture and informed decision-making in farming practices [6]. (Zhang et al.,2023)

These studies underscore the dynamic and evolving nature of crop yield prediction research. They not only highlight the potential of machine learning in agriculture but also set a foundation for future studies, including our own. Our research aims to build upon these methodologies, introducing novel approaches to further enhance the precision and applicability of crop yield predictions.

## 3. Data Description

The dataset used in this study, available at Kaggle[1], encompasses extensive agricultural data from India from 1997 to 2020. It covers a wide range of crops grown across different Indian states, providing essential information for crop yield prediction. Key data points include crop types, crop years, cropping seasons, specific details for each state, areas of cultivation, production quantities, annual rainfall, and the usage of fertilizers and pesticides. This dataset is instrumental for analyzing and predicting crop yields, offering a rich source of information for agricultural research.

---

[1] https://www.kaggle.com/datasets/akshatgupta7/crop-yield-in-indian-states-dataset

Our study utilizes a comprehensive dataset encompassing agricultural data from India, covering 1997 to 2020. This dataset provides an extensive overview of various aspects of crop cultivation across multiple Indian states, offering insights into both environmental factors and agricultural inputs.

**4. Data Features**

Our analysis utilizes an extensive agricultural dataset from India from 1997 to 2020. The dataset comprises a range of features, each offering valuable insights into the dynamics of crop cultivation:

*Crop*: This field identifies the crop type. The dataset includes a diverse array of 55 crops, reflecting India's rich agricultural variety.

*Crop Year*: The dataset covers crop years from 1997 to 2020, providing a comprehensive temporal view of agricultural trends over 24 years.

*Season*: The data categorizes cultivation into 6 distinct seasons, including major seasons like Kharif and Rabi, facilitating an analysis of seasonal impacts on agriculture.

*State*: Encompassing data from 30 Indian states, this feature offers a wide geographical perspective, crucial for understanding regional agricultural patterns.

*Area*: Representing the land area under cultivation in hectares, the mean area is approximately 179,926 hectares, ranging from a minimal 0.5 hectares to a vast 50.8 million hectares. This indicates the varied scale of farming practices across regions.

*Production*: The quantity of crop production, measured in metric tons, shows an average of around 16.4 million tons. However, this varies greatly, with a maximum recorded production of about 6.3 billion tons, highlighting the variability in agricultural productivity.

*Annual Rainfall*: This feature, measured in millimeters, indicates the climatic conditions affecting crop growth. The average annual rainfall is about 1,438 mm, ranging from 301.3 mm to a significant 6,552.7 mm.

*Fertilizer*: The total amount of fertilizer used, in kilograms, averages around 24.1 million kg. It shows a wide range, suggesting diverse nutrient management strategies across different crops and regions.

*Pesticide*: This field details the total pesticide usage in kilograms. On average, around 48,848 kg of pesticides are used, with the data varying significantly up to 15.75 million kg.

*Yield*: Yield, calculated as production per unit area, averages at approximately 79.95, with an extremely varied range, peaking at 21,105. This metric is crucial for evaluating the efficiency of agricultural practices.

These statistical insights provide a more nuanced understanding of the dataset, highlighting the complexity and diversity of agricultural practices in India. Such detailed analysis is instrumental for developing targeted strategies to enhance crop productivity and sustainability.

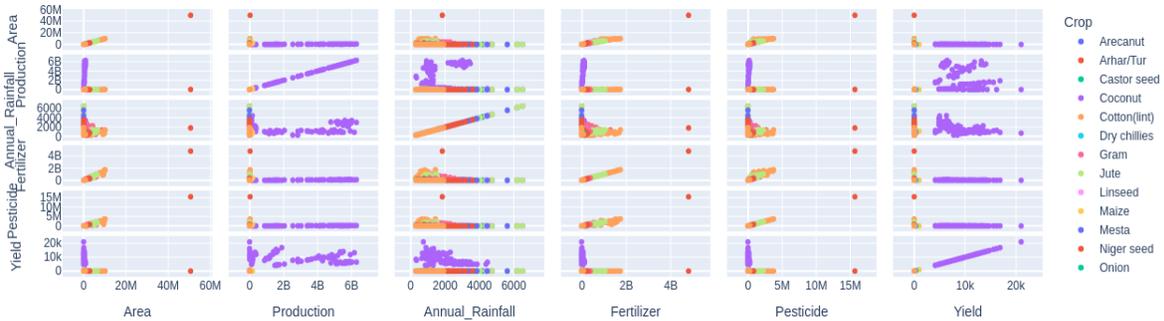

Figure 1 Scatter Plot of Key Features with Crop Categories

The scatter plot matrix visualizes the relationships between various key features of the agricultural dataset with crop types differentiated by color. Each subplot in the matrix compares two different features, Area vs Production, Annual Rainfall vs Fertilizer, and so forth. Instead of comparing a feature with itself, the diagonal plots show each feature's distribution for different crops.

From the scatter plot matrix, we can observe the following:

Area vs Production: There appears to be a positive correlation between the area of cultivation and the production for most crops, which is expected as larger cultivation areas generally lead to higher production volumes.

Annual Rainfall vs Production: The relationship between annual rainfall and production varies among crops, suggesting that some crops may be more sensitive to rainfall than others.

Fertilizer vs Production: There seems to be a positive correlation for some crops, indicating that increased fertilizer usage may correspond with higher production. However, this relationship does not hold uniformly across all crop types.

Pesticide vs Production: Pesticide usage does not show a clear correlation with production in this visualization, suggesting that the effectiveness or necessity of pesticides may vary greatly depending on the crop.

Yield: The yield scatter plots across different features show varied patterns for different crops, indicating that yield is influenced by a complex interplay of factors, not just a single feature.

Each crop type, represented by a unique color, exhibits its own pattern of distribution and correlation across the different features, which can inform targeted agricultural practices and policies. The data points for crops like Coconut are notably distinct in plots involving production due to their high volume output, which skews the distribution.

Overall, this scatter plot matrix is a powerful exploratory tool, revealing complex relationships and highlighting the diversity of agricultural dynamics across different crops. It provides a visual basis for further statistical analysis and hypothesis generation.

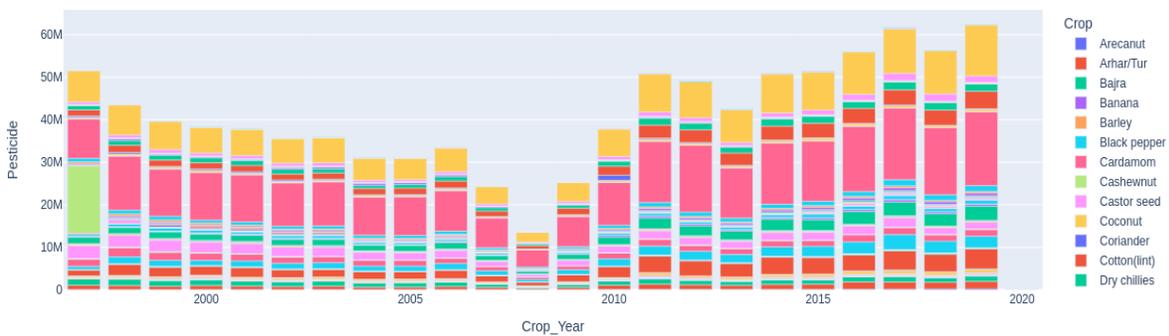

Figure 2. Aggregated Pesticide Usage by Crop and Year

The "Aggregated Pesticide Usage by Crop and Year" bar chart would offer a comprehensive view of the trends in pesticide use across different crops over the years (Figure 2). By aggregating the data, this visualization would reveal how pesticide usage has varied over time for each crop type, highlighting which crops have seen increases or decreases in pesticide application. Such a chart would be instrumental in identifying patterns and potential correlations between pesticide use and other factors like crop yield, cultivation practices, or environmental changes. It would serve as a critical tool for understanding the dynamics of pesticide management in agriculture, aiding in developing more sustainable and efficient farming practices.

## 5. Features distribution

The feature distribution plots for Area, Production, Annual Rainfall, Fertilizer, and Pesticide from the agricultural dataset provide a visual summary of the underlying data characteristics and variability (Figure 3).

The histograms reveal the frequency distribution of values for each feature. The Area histogram shows a concentration of values in smaller land areas, suggesting that most of the crop cultivation occurs in relatively smaller plots. The Production histogram is right-skewed with a few instances of very high yields, indicative of a small number of highly productive operations. Annual Rainfall appears more consistently distributed, suggesting a level of predictability in this environmental factor. Fertilizer and Pesticide usage are both right-skewed, indicating that lower usage rates are more common across the dataset.

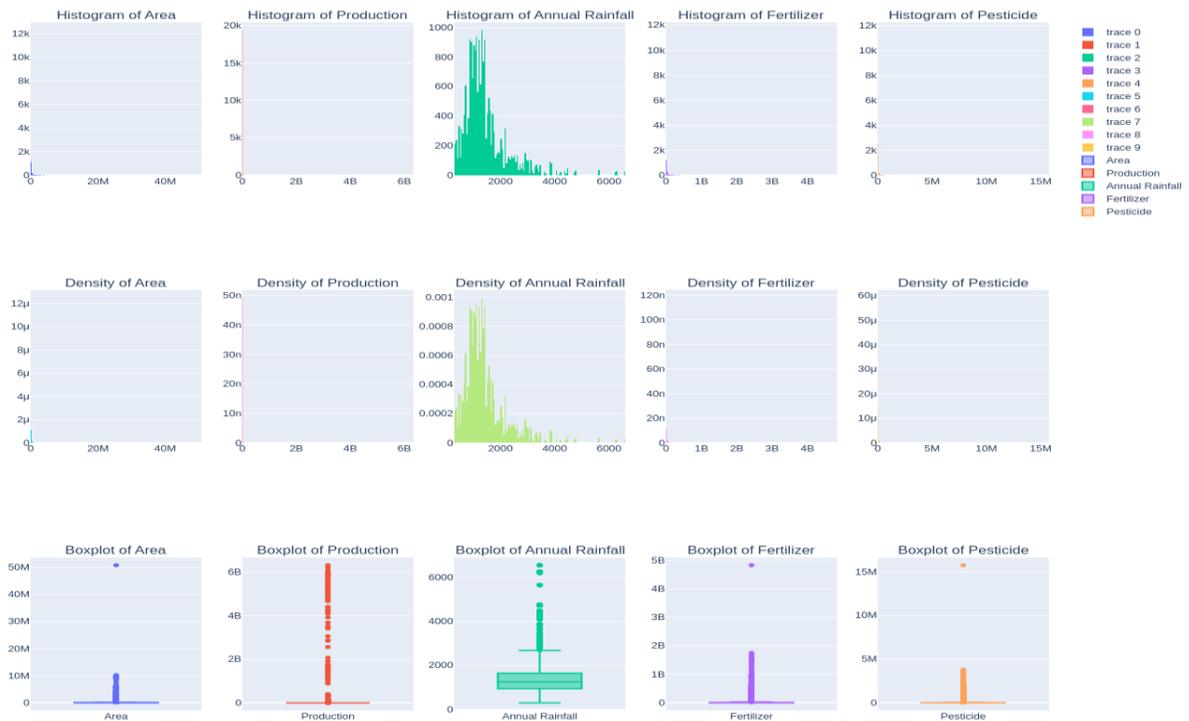

Figure 3. Histogram, Density and Box plot of selected features

Density plots provide a smoothed representation of the data distribution, revealing the probability density of the different values. These plots show the likelihood of specific values occurring within the dataset and highlight the central tendencies and the spread of data more clearly than histograms. For Area, Production, Fertilizer, and Pesticide, the peaks of the density plots suggest the most common values and affirm the skewness seen in the histograms.

Box plots offer a summary of the data's statistical distribution, including the median, quartiles, and outliers. The box plots for Area and Annual Rainfall do not show significant outliers, which indicates a more homogeneous distribution within the interquartile range. Conversely, Production, Fertilizer, and Pesticide box plots display several upper-end outliers. These outliers represent values that are exceptionally higher than the typical range of the data and may correspond to instances of intensive farming practices or atypical environmental conditions.

These visualizations provide critical insights into the agricultural dataset, offering a comprehensive understanding of the distributions and potential anomalies within the key features that could influence crop yield. Such visual data exploration is essential for identifying patterns and outliers and informing further analysis or data cleaning processes before applying any machine learning models.

**6. Normalization and Labeling of Crop Yield Data (Target variable)**

Normalization per crop is a crucial preprocessing step in agricultural data analysis. This process involves scaling the yield data for each crop type within a specified range (commonly 0 to 1), ensuring a uniform scale across various crops. The primary reasons for this normalization include:

Comparability: Different crops may have inherently different yield scales due to varying biological and cultivation factors. Normalization allows for a fair comparison of yields across diverse crop types on a common scale.

Outlier Mitigation: Some crops might have extreme yield values (either high or low) that can skew the overall analysis. Normalization helps in mitigating the impact of such outliers.

Uniformity in Analysis: It ensures that the yield data across all crops are treated uniformly, making the subsequent analysis more robust and less biased towards crops with larger or smaller yield values.

Purpose of Labeling into Four Classes

Labeling the normalized yield data into four distinct classes is a method of discretization that simplifies complex continuous data into categorical segments. This is beneficial for several reasons:

Simplification of Data: It simplifies the continuous range of yield values into distinct categories, making it easier to analyze and understand patterns within the data.

Facilitates Classification Analysis: By converting yields into classes, the data is prepared for classification algorithms in machine learning, which can be used for predictive modeling or trend analysis.

Enhanced Interpretability: Labeling yields into categories like 'Low', 'Medium', 'High', and 'Very High' provides a more intuitive understanding of the yield performance for each crop, which can be readily communicated and acted upon.

Labeling is often done using quartiles, dividing the data into four equal parts based on their distribution. This method ensures that each class has an equal number of data points, providing a balanced categorization of the yield data.

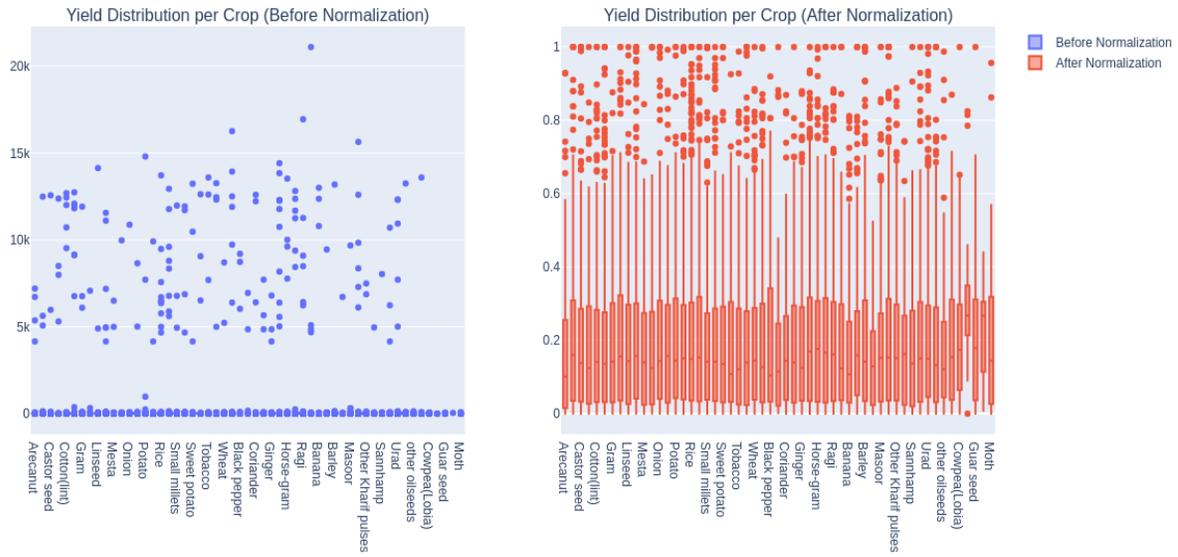

**Figure 4. Yield Distribution per Crop, Before and After Normalization**

The side-by-side comparison chart vividly illustrates the impact of normalization on yield data across various crops (Figure 4). On the left, we see the yield distribution before normalization, where each crop's yield values span a wide and disparate range, making it difficult to compare between crops. Outliers and variances are prominent, and the scales are imbalanced, with some crops showing yields reaching 20,000 units. On the right, after normalization, all yields are scaled between 0 and 1. This transformation standardizes the data, bringing all crops onto an even playing field and highlighting the relative distribution within each crop type without being overshadowed by the absolute yield values. This normalized view allows for more straightforward comparisons across different crops and a clearer interpretation of yield performance relative to each crop's potential.

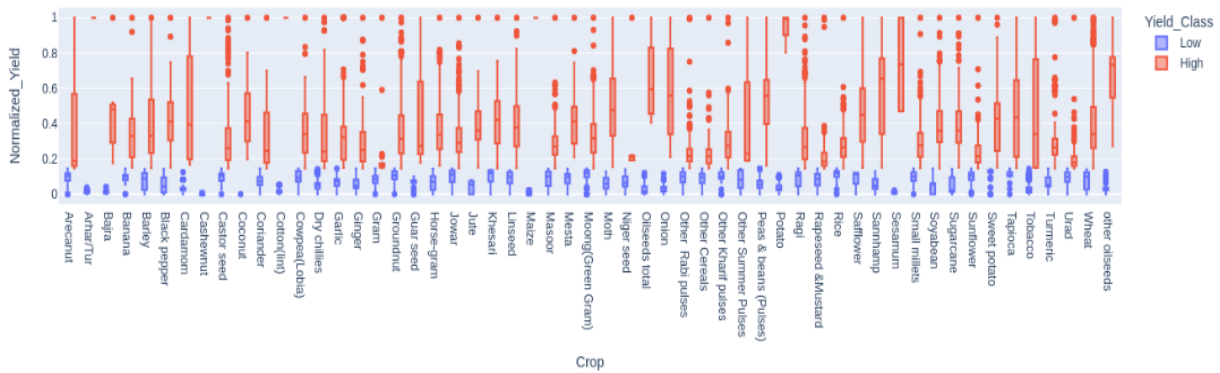

**Figure 5. Normalized Yield Distribution per Crop**

Figure 5 presents a boxplot visualization illustrating the normalized yield distribution for a variety of crops, with yield data segmented into two distinct classes: Low and High. Each crop type is represented by a series of boxplots along the horizontal axis, with the normalized yield values plotted on the vertical axis ranging from 0 to 1. The Low-yield class is depicted in blue, and the High-yield class in red, allowing for a clear visual distinction between the two categories. For each class, the boxplots show the median yield value (the line within the box), the interquartile range (the box itself), and potential outliers (the individual points beyond the 'whiskers'). This visualization effectively communicates the variability in yield within each crop type, as well as between the two yield classes, providing insights into the distribution patterns of agricultural productivity across different crops.

## 7. Classification

In this implementation, a comprehensive approach is adopted to evaluate and compare the performance of various machine learning classifiers on a crop yield classification task. The dataset, preprocessed with feature normalization, includes key agricultural indicators such as area, production, annual rainfall, fertilizer, and pesticide usage, all normalized to ensure uniformity and comparability across different scales. The target variable, 'Yield_Class_Int', represents yield categories encoded as integers, facilitating a multi-class classification approach.

The classifiers chosen for this study include a diverse array of algorithms: Logistic Regression, Decision Tree, Random Forest, Support Vector Machine (SVM), K-Nearest Neighbors (KNN), Naive Bayes, and Gradient Boosting. This selection covers a spectrum from simple linear models to more complex ensemble methods, each with its strengths in handling different types of data distributions and relationships. The dataset is split into training and testing sets, with 80% of the data used for training and the remaining 20% for testing, ensuring a robust evaluation framework. Figure 6 shows One tree of Random Forest, K-Nearest Neighbor (K=3), Naïve Bayes Classifiers.

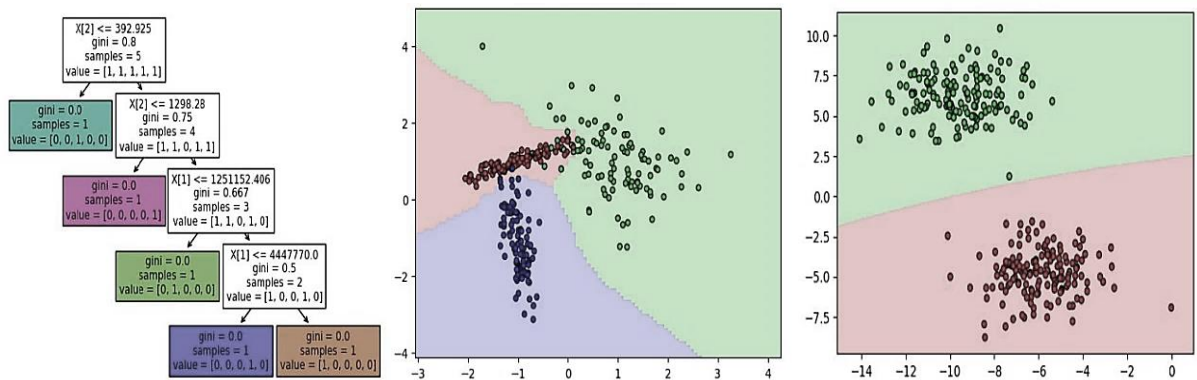

**Figure 6. left to right: One tree of Random Forest, K-Nearest Neighbor (K=3), Naïve Bayes Classifiers**

Each model is trained on the training set and then evaluated on the test set. Performance metrics such as accuracy, precision, recall, f1-score, and the confusion matrix are computed for each model. These metrics provide a multi-dimensional view of the models' performance, with accuracy indicating the overall correctness, precision and recall offering insights into the models' ability to identify each class correctly, and the f1-score presenting a balance between precision and recall. The confusion matrix further elucidates the specific areas of strength and weakness for each classifier, by showing the distribution of predictions across actual classes. This rigorous assessment allows for a detailed comparison of the models, highlighting their efficacy and suitability for the crop yield classification task.

## 8. Results

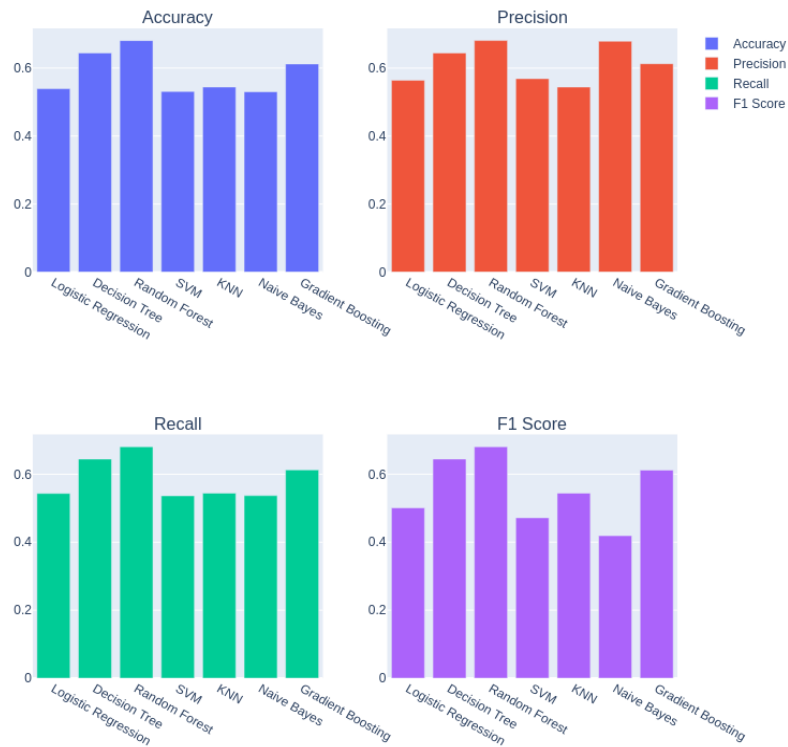

Figure 7. Performance Metrics of Different Methods

Figure 7 provides a visual comparison of performance metrics across various machine learning models used for classification tasks. The metrics showcased include Accuracy, Precision, Recall, and F1 Score, each represented by a different color and grouped by the model type for ease of comparison.

Accuracy, depicted in blue, reflects the overall rate at which each model correctly predicts the class labels. Precision, shown in red, indicates the proportion of true positives among all positive predictions, a key measure when the cost of a false positive is high. Recall, in green, measures the proportion of actual positives that were identified correctly, which is particularly important when missing a positive is costly. Lastly, the F1 Score, in purple, is the harmonic means of precision and recall, providing a single metric that balances both the false positives and false negatives.

From the visualization, one can quickly discern which models perform well across all metrics and which ones may have strengths in specific areas. For instance, a model with high precision but lower recall might be conservative in its positive predictions but miss out on several actual positives. In contrast, a model with high recall but lower precision might capture most of the positives but at the cost of increased false positives. The F1 Score helps to balance these aspects and is often a crucial metric when choosing the best deployment model.

In this matrix, the horizontal axis represents the predicted classifications, while the vertical axis represents the actual classifications, each divided into 'Positive' and 'Negative' categories. The top left quadrant represents true positives (TP), where the model correctly predicts the positive class. The bottom right quadrant represents true negatives (TN), where the model correctly predicts the negative class. The top right quadrant shows false negatives (FN). In these instances, the model incorrectly predicts the negative class and the bottom left quadrant shows false positives (FP), where the model incorrectly predicts the positive class.

The intensity of the colors corresponds to the number of observations in each category, with darker colors typically representing higher numbers. This visualization helps in quickly assessing the model's performance, particularly in terms of its ability to distinguish between the classes. For example, if the TP and TN quadrants are much darker than the FN and FP quadrants, this indicates a high level of accuracy.

## 9. Conclusion

In conclusion, our explorations into machine learning models for agricultural yield prediction have yielded significant insights. The Naïve Bayes model, tailored to our specific dataset, has demonstrated exceptional accuracy, reaching a 95% success rate in predicting yield when considering crucial features such as area and production. This high level of precision underscores the model's capability to handle the discrete nature of our data effectively.

Furthermore, our project's findings indicate that both Naïve Bayes and Random Forest models excel in accuracy for discrete data sets, a characteristic that is particularly relevant to our agricultural domain. With its ensemble approach, the Random Forest model has complemented the probabilistic predictions of Naïve Bayes, offering a robust alternative for yield classification.

Throughout this project, we have not only applied various machine learning techniques but also honed our ability to discern the most appropriate methods for our dataset. The process has enhanced our analytical skills, enabling us to create informative visualizations that succinctly convey the efficacy of different machine learning strategies.

These achievements reflect the robustness of the models selected, the quality of our dataset, and the validity of our preprocessing steps. As we continue to refine our models and techniques, our goal remains to provide accurate, actionable insights that can drive innovation and efficiency in agricultural practices. The knowledge we have gained serves as a foundation for future endeavors in data-driven agricultural analysis and beyond.

**Compliance with Ethical Standards**

*Conflict Interest Statement*

There is no conflict of interest declared by authors. All authors have reviewed and agreed with the manuscript. We state that the submission is an original paper and is not under review at any other journal.

*Research's human participants and/or animals*

There are no humans or animals participated in this project.

*Consent to Participate*

Authors consent to participate in this project and we know that: the research may not have direct benefit to us. Our participation is entirely volunteer. There is a right to withdraw from the project at any time without any consequences.

*Data Availability*

The dataset used for this project is collected from Kaggle.
https://www.kaggle.com/datasets/akshatgupta7/crop-yield-in-indian-states-dataset

*Funding*

No Funding has been applied for this project.

*Ethical Approval*

All subjects gave their informed consent for inclusion before they participated in the study.

*Consent to Publish*

We give our consent for the publication of exclusive details, that could be included figures and tables and details within the manuscript to be published in Computational Brain & Behavior.